\pdfoutput=1

\documentclass[10pt,twocolumn,letterpaper]{article}

\usepackage[pagenumbers]{cvpr}

\usepackage{bm}
\usepackage{tikz}
\usepackage{float}
\usetikzlibrary{arrows.meta,calc}
\IfFileExists{microtype.sty}{\usepackage{microtype}}{}

\usepackage{cuted}

\usepackage{currfile}

\usepackage{caption}

\newcommand{\method}{Spaghetti Splatting\xspace}
\newcommand{\vA}{\bm{a}}
\newcommand{\vB}{\bm{b}}
\newcommand{\vx}{\bm{x}}
\newcommand{\vv}{\bm{v}}
\newcommand{\vo}{\bm{o}}
\newcommand{\vd}{\bm{d}}
\newcommand{\vmu}{\bm{\mu}}
\newcommand{\mQ}{\bm{Q}}
\newcommand{\mS}{\bm{\Sigma}}
\newcommand{\gaussian}{\mathcal{G}}
\newcommand{\conv}{\circledast}
\newcommand{\dd}{\mathrm{d}}

\usepackage[ruled]{algorithm2e}
\SetAlFnt{\small}
\SetAlCapFnt{\small}
\SetAlCapNameFnt{\small}
\SetAlCapHSkip{0pt}

\AtEndPreamble{\Crefname{section}{Section}{Sections}
\crefname{section}{Sec.}{Secs.}
\Crefname{appendix}{Appendix}{Appendices}
\crefname{appendix}{App.}{Apps.}
\Crefname{table}{Table}{Tables}
\crefname{table}{Tab.}{Tabs.}
\Crefname{figure}{Figure}{Figures}
\crefname{figure}{Fig.}{Figs.}
\Crefname{equation}{Eq.}{Eqs.}
\crefname{equation}{}{}
}

\usepackage{soul}
\setuldepth{foobar}

\usepackage{pifont}

\definecolor{dark_green}{rgb}{0, 0.5, 0}

\newcommand{\prim}{\mathcal{P}}
\newcommand{\SupplementaryMaterial}{supplementary material}

\newcommand{\calN}{\mathcal{N}}

\usepackage{colortbl}
\usepackage{booktabs}

\definecolor{medalgold}{RGB}{221,196,65}
\definecolor{medalsilver}{RGB}{215,215,215}
\definecolor{medalbronze}{RGB}{126,66,5}
\newcommand{\medal}[1]{\tikz[baseline=-0.7ex]\draw[#1,fill=#1,radius=2pt](0,0)circle;}
\newcommand{\mgold}[1]{\textbf{#1}\rlap{ \medal{medalgold}}}
\newcommand{\msilver}[1]{#1\rlap{ \medal{medalsilver}}}
\newcommand{\mbronze}[1]{#1\rlap{ \medal{medalbronze}}}

\usepackage{multirow}
\usepackage{booktabs,graphicx,array}

\usepackage{array}
\usepackage{booktabs}
\usepackage{multirow}
\usepackage{tabularx}

\renewcommand{\paragraph}[1]{\vspace{.25em}\noindent\textbf{#1.}}

\definecolor{cvprblue}{rgb}{0.21,0.49,0.74}
\usepackage[pagebackref,breaklinks,colorlinks,allcolors=cvprblue]{hyperref}

\def\confName{CVPR}
\def\confYear{2027}

\title{{\LARGE \method} \\ Differentiating Curvilinear Dirac Masses via Convolution}

\author{
George Shramko$^{1}$ \quad
Shrisudhan Govindarajan$^{1,2}$ \quad
Rudy Lam$^{1}$ \quad
Yizheng Xie$^{1}$
\\ 
Daniel Rebain$^{2,3}$ \quad
David B. Lindell$^{4}$ \quad
Andrea Tagliasacchi$^{1,2,4}$
\\[0.25cm]
\normalfont\normalsize
$^{1}$Simon Fraser University \quad
$^{2}$Wayve Labs \quad
$^{3}$University of British Columbia \quad
$^{4}$University of Toronto}

\begin{document}
\maketitle
\begin{strip}
\vspace{-2em}
\centering
\setlength{\abovecaptionskip}{3pt}\includegraphics[width=\textwidth]{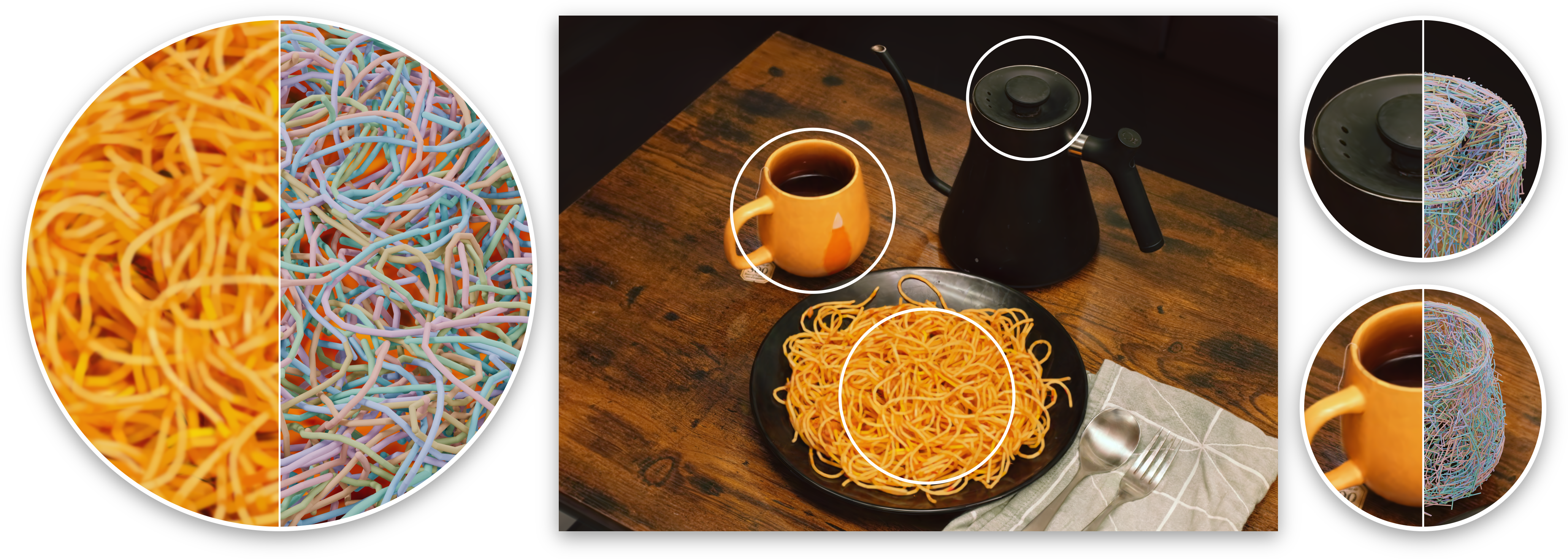}
\captionsetup{hypcap=false}
\vspace{0em}
\captionof{figure}[Teaser.]{
\textbf{Teaser} --
We introduce a differentiable 3D representation which can be used to learn accurate radiance fields for novel view synthesis applications~(left).
We showcase the internal structure of our representation~(right) by coloring each primitive with unique colors.
\footnotemark
This shows that our primitives are a set of connected polylines~(i.e. \textit{spaghetti}), each obtained by convolving a polyline Dirac measure with a Gaussian or beta kernel.
It is \emph{splatting}, as each primitive is rendered by splatting its kernel onto the image.\footnotemark
}
\captionsetup{hypcap=true}
\label{fig:\currfilebase}\end{strip}
\makeatletter
\addtocounter{footnote}{-1}\edef\Hy@footnote@currentHref{Hfootnote.\the\numexpr\value{Hfootnote}-1\relax}\footnotetext{We thank Zhen~(Colin) Li for help with capturing the pasta scene, and Francesco Di Sario and Maria Taktasheva for cooking, as well as eating the pasta.}\stepcounter{footnote}\edef\Hy@footnote@currentHref{Hfootnote.\the\value{Hfootnote}}\makeatother
\footnotetext{We provide details on how we perform the reconstruction for this teaser in the \SupplementaryMaterial{}.
}

\begin{abstract}
In production systems, 3D geometry is represented by primitives such as points, lines, and triangles, as rendering hardware is optimized to generate images from them very efficiently.
Learning such primitives from images by volume rendering is challenging, as they are Dirac masses defined at infinitesimal locations in space.
EWA splatting, Poisson reconstruction, and 3D Gaussian Splatting overcome this by convolving Dirac masses with a low-frequency kernel, a recipe that generalizes along two axes: the kernel and the primitive.
As points and triangles have already received significant attention, we investigate this recipe for \emph{curvilinear} primitives, and introduce \method.
We make curves differentiable by convolving them with a kernel, and show that the convolution of a polyline converges to that of the continuous curve as its segments shrink.
We approximate curves by polylines, and volume render each segment so that connected segments composite correctly.
For fast training and inference, we linearize the camera projection per segment, which accommodates kernels beside Gaussians, such as Beta kernels.
As a point is a curve of zero length, our representation is a strict superset of 3D Gaussian Splatting.
Hence, it remains competitive with the state-of-the-art on general scenes, outperforms methods specialized to line-like geometry, natively supports anti-aliasing, and produces long connected strands that are directly amenable to physics simulation.
\end{abstract}

\begin{figure}[h]
\centering
\includegraphics[width=\linewidth]{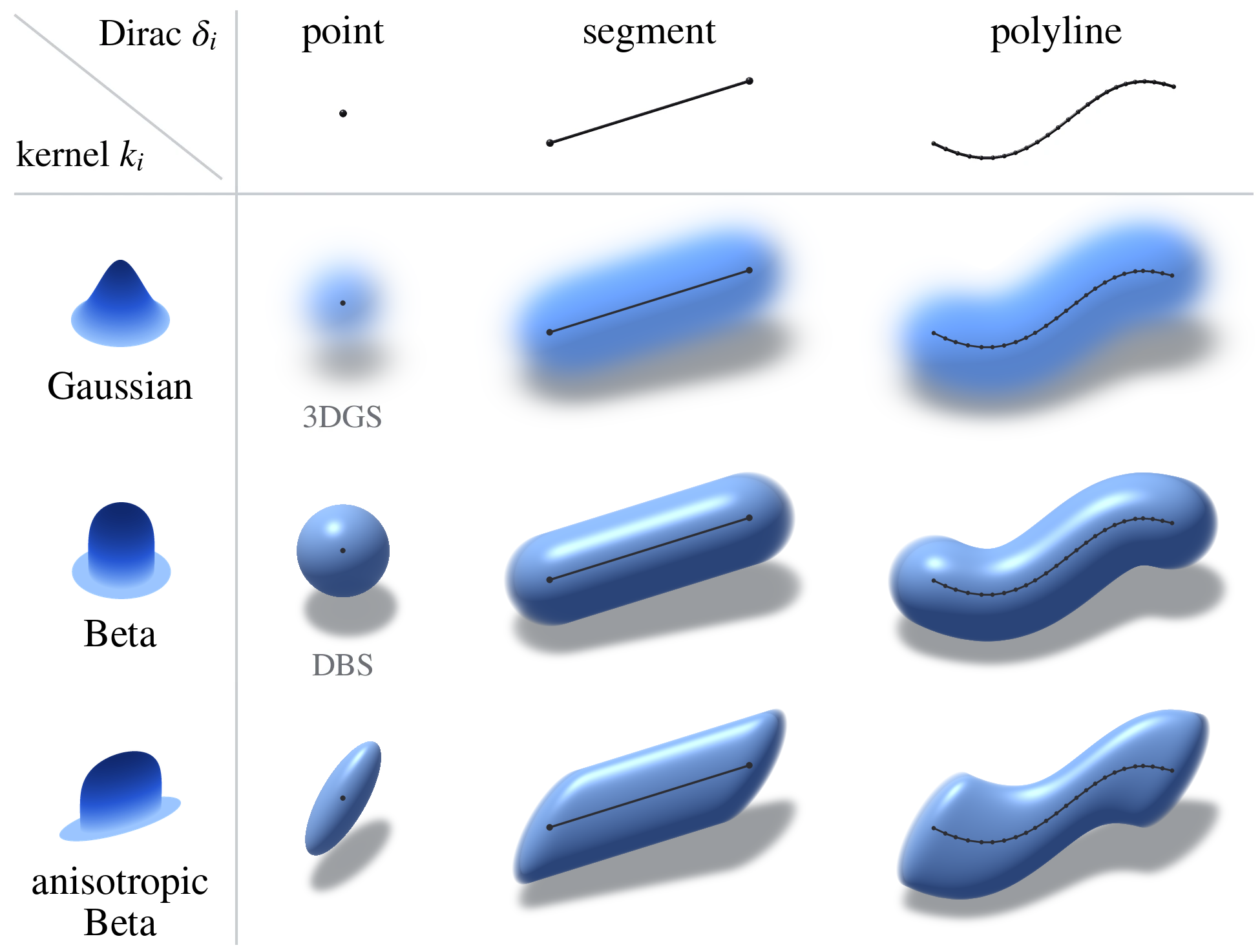}
\caption{
\textbf{Kernel $\conv$ dirac} --
Since Dirac masses and convolution kernels are independent degrees of freedom, we can modify either one to obtain a different differentiable primitive.
The first column shows that convolving a point mass with a Gaussian or a Beta kernel yields the primitives of 3DGS~\cite{kerbl2023gaussians} and DBS~\cite{liu2025dbs}, respectively.
Replacing the point mass with a line segment, while retaining either kernel, yields the primitives in the middle.
Finally, we connect these segments into a polyline, which yields our \emph{curvilinear} primitive.
In the bottom row, we convolve the same Dirac masses with an \textit{anisotropic} Beta kernel.
For visualization, we shade the primitives with the normals of their convolution surfaces~\cite{bloomenthal1991convolution}.
}
\label{fig:\currfilebase}\end{figure}

\begin{figure*}[t]
\centering
\includegraphics[width=\textwidth]{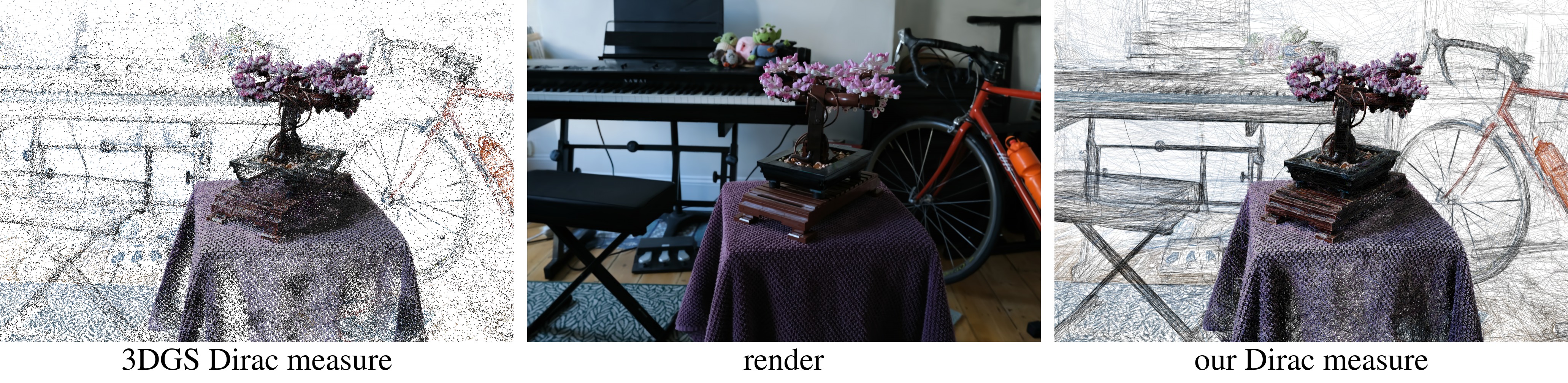}
\caption{
\textbf{Inside our spaghetti} -- 
We visualize the Dirac measure of~\citet{kheradmand2024mcmc}~(left), and of our \method~(right) for the Bonsai scene from Mip-NeRF 360~\cite{barron2022mipnerf360}.
(middle) We show one RGB render, not repeated, as it is visually identical for both representations.
While 3DGS models thin strokes with scattered points (e.g. the bicycle spokes), our segments form continuous strokes that follow edges.
}
\label{fig:\currfilebase}
\end{figure*}

\section{Introduction}
\label{sec:intro}

In production graphics systems, 3D geometry is represented by points, lines, and triangles~(\ie, \emph{primitives}), as rendering hardware is optimized to generate images from them very efficiently.
Rather than constructing them by hand, we would like to \emph{optimize} such primitives directly from a collection of images, by volume rendering~\cite{mildenhall2020nerf} and gradient descent.
However, this is challenging, as primitives are Dirac masses defined at \textit{infinitesimal} locations in space, so there is nothing for a gradient to act on.

Research from about two decades ago showed us how to overcome this.
EWA splatting~\cite{zwicker2002ewa} and Poisson reconstruction~\cite{kazhdan2006poisson} both turn a Dirac-like representation into a volumetric one by \emph{convolving} it with a low-frequency kernel.
This makes sampling the representation onto a discrete grid of pixels well posed from the perspective of the Nyquist--Shannon theorem~\cite{yu2024mipsplatting}.
3D Gaussian Splatting~\cite{kerbl2023gaussians} uses this idea to differentiate through point-like primitives, which it convolves with anisotropic Gaussian kernels.
\textit{But could this idea be generalized?}

There are two potential axes of generalization, made apparent by the $\mathcal{L}_2$ objective that these methods optimize:
\begin{equation}
  \mathcal{L}_2=\sum\nolimits_{c}\Big\lVert\bm I_c-\mathcal{R}_c\Big(\sum\nolimits_i\delta_i\conv k_i\Big)\Big\rVert_2,
  \label{eq:convolution}
\end{equation}
where $\bm I_c$ is the image taken from camera viewpoint $c$, $\mathcal{R}_c$ volume renders the scene from that same viewpoint, and $\delta_i$ is the Dirac measure of the $i$-th primitive, convolved with kernel $k_i$.
One axis generalizes the \textit{kernel}, as in Deformable Beta Splatting~\cite{liu2025dbs}.
The other generalizes the \textit{primitive}, with initial research on convexes~\cite{held2025convex}, triangles~\cite{held2026triangles}, and, very recently, lines~\cite{tojo2026lines}.\footnote{Note that the work by \citet{tojo2026lines} is \ul{concurrent} to ours.}
As points~\cite{kerbl2023gaussians,yu2024mipsplatting,liu2025dbs} and triangles~\cite{held2026triangles,held2025meshsplatting} have already received significant attention, we focus on differentiating through \emph{curvilinear} primitives; see \Cref{fig:convolution}.

To this end, we approximate curves by polylines, and show that their convolution with a kernel converges to that of the continuous curve as the segments shrink.
Whereas the Gaussian convolution of a polygon requires evaluating Owen's T function~\cite{owen1956tables,didonato1980polygons}, the ray integral of a Gaussian-convolved segment only involves the Gauss error function~$\operatorname{erf}$~(\ie, the anti-derivative of a Gaussian).
By volume rendering the density of each segment~\cite{tagliasacchi2022volumerenderingdigestfor}, connected segments composite without overcounting at their joints.
For fast training and inference, we follow EWA splatting~\cite{zwicker2002ewa} and linearize the camera projection per segment, which also accommodates Beta kernels~\cite{liu2025dbs}.

Our representation is a strict superset of 3D Gaussian Splatting, as a point is nothing else than a curve of zero length.
As such, it matches or outperforms the state-of-the-art on Mip-NeRF~360~\cite{barron2022mipnerf360}, \ie, Deformable Beta Splatting~\cite{liu2025dbs} for geometry, coupled with Spherical Voronoi~\cite{disario2026sv} for modeling of view-dependent appearance.
\method also outperforms methods specialized to line-like geometry: the concurrent work of \citet{tojo2026lines} on their Fuzzy benchmark, and Gaussian Haircut~\cite{zakharov2024haircut}, which relies on hair-specific priors, on GeomHair~\cite{lazuardi2026geomhair}.
It natively supports Mip-Splatting~\cite{yu2024mipsplatting} anti-aliasing, so it remains stable across viewing distances.
Finally, our representation consists of long connected strands, which makes it amenable to physics simulation, and closer to what an artist would produce.

\vspace{.5em}
\noindent
We achieve these results through four key contributions:
\begin{itemize}
\item We note that the key ingredient behind 3D Gaussian Splatting and Poisson reconstruction, \ie, convolving Dirac masses with a low-frequency kernel, provides a recipe that generalizes along both the kernel and the primitive.
\item We investigate this recipe for \emph{curvilinear} Dirac masses, and provide a convergent as well as efficient way to back-propagate through them.
\item We generalize 3D Gaussian Splatting to a strictly larger family of representations, which remains competitive with the state-of-the-art on general scenes, while setting a new state-of-the-art for strand-based representations.
\item We commit to \ul{fully release code} to foster precise comparisons, and follow-up research.
\end{itemize}

\section{Related work}
\label{sec:related}
Neural Radiance Fields~(NeRF) represents a scene as a continuous density and color field rendered by numerical integration along camera rays~\cite{mildenhall2020nerf}.
Later neural fields addressed aliasing and unbounded scenes~\cite{barron2021mipnerf,barron2022mipnerf360,barron2023zipnerf}, while multiresolution hash encodings substantially reduced their training cost~\cite{mueller2022instant}.
3DGS replaced the neural field with explicit anisotropic Gaussians~\cite{kerbl2023gaussians}, and a tile-based rasterizer derived from point-based volume rendering and EWA splatting~\cite{zwicker2002ewa,westover1990footprint}.
Subsequent methods have improved both its optimization and rendering.
3DGS-MCMC formulates population control as sampling~\cite{kheradmand2024mcmc}, and StopThePop introduces per-pixel sorting for more consistent visibility~\cite{radl2024stopthepop}.
Other methods parallelize the backward pass over Gaussians instead of pixels~\cite{mallick2024taming}, integrate Gaussian density along each ray~\cite{talegaonkar2025vol3dgs}, render constant-density ellipsoids in closed form~\cite{mai2025ever}, or avoid the affine projection used by rasterization~\cite{moenneloccoz2024threedgrt,wu2025threedgut}.
Another line of work replaces the Gaussian kernel: Deformable Beta Splatting~(DBS) learns the fall-off of a compact Beta kernel~\cite{liu2025dbs}, while Student Splatting and Scooping uses heavy-tailed kernels with signed weights~\cite{zhu2025sss}.
Mip-Splatting prevents aliasing across viewing distances with a three-dimensional smoothing filter based on the training views and a two-dimensional filter during rasterization~\cite{yu2024mipsplatting}.
Spherical Voronoi replaces spherical harmonics with a learned directional partition~\cite{disario2026sv}.
We build on several of these improvements and adopt 3DGS-MCMC population control, the Beta kernel of DBS, Mip-Splatting filtering, and Spherical Voronoi appearance modeling.

\begin{figure*}[t]
  \centering
  \includegraphics[width=\textwidth]{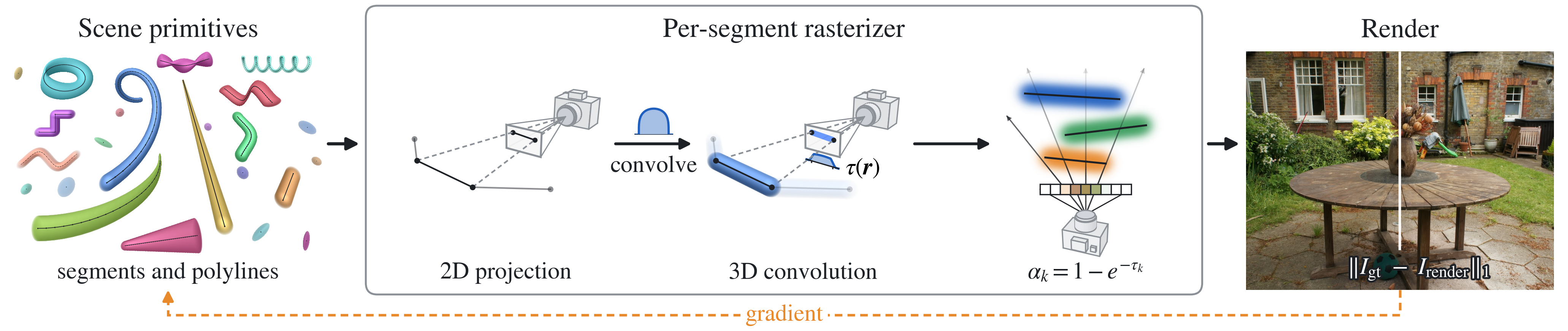}
  \caption{
  \textbf{Overview} --
  (Left) Our representation is a set of learnable polyline primitives convolved with Gaussian or Beta kernels.
  (Middle) Without loss of generality, we render each segment individually: we first project it onto the image plane, and then evaluate the integral of the convolution along the ray in screen space using the closed-form integrals of \Cref{sec:exact}, which yields a per-segment screen-space rendering.
  The segments are then composited along each ray to obtain the accumulated color.
  (Right) We compute the $L_1$ loss between the rendered image and the ground-truth image, and backpropagate its gradient to the parameters of the primitives.
  }
  \label{fig:\currfilebase}
\end{figure*}

\paragraph{Beyond point primitives}
Another line of work replaces point centers with extended geometric primitives.
Triangle Splatting~\cite{held2026triangles} and 3D Convex Splatting~\cite{held2025convex} optimize triangles and smooth convex bodies.
Curve primitives have so far targeted sparse abstractions of a scene: 3Doodle optimizes 3D B\'ezier strokes and superquadric contours to render view-consistent sketches~\cite{choi2024doodle}, while SketchSplat reconstructs parametric 3D edges by splatting Gaussians sampled along curves onto multi-view edge maps~\cite{ying2025sketchsplat}.
The concurrently developed \ul{opaque} line primitives of \citet{tojo2026lines} are closely related to our work, as they also use line primitives directly for differentiable rendering.
Their method rasterizes segments on a subpixel grid using the Bresenham algorithm and recovers high-quality geometry in object-centric scenes.
That said, their line primitives are rendered at a constant screen-space width regardless of their depth from the camera, making the representation unsuitable for general scenes, as we show in our comparison on Mip-NeRF~360~(\cref{fig:benchmarks}).
We include evaluations of our method on the Fuzzy dataset introduced by their work.

\paragraph{Convolution surfaces}
A classical computer graphics technique, Convolution Surfaces~\cite{bloomenthal1991convolution}, has shown that one can represent many implicit shapes as an iso-surface of a convolution of a skeleton of curves and polygons with a smooth kernel.
Recently, \citet{xu2025gpis} demonstrated that one can efficiently render a Gaussian process implicit surface as an iso-surface of a sparse convolution of a kernel with randomly placed points.
In both cases, the convolved field serves only to define an iso-surface.
We instead use segment and polyline skeletons convolved with corresponding kernels as an explicit volume-rendering representation which we optimize from multi-view images.
In this view, point-based splatting primitives are a special case of our segments with zero length, and our method maintains their fast rendering speed.

\paragraph{Hair reconstruction}
Hair is a prominent example of geometry that is naturally described by curves.
Neural Strands and Neural Haircut reconstruct strand geometry with learned priors~\cite{rosu2022neuralstrands,sklyarova2023neuralhaircut}.
Gaussian Haircut and GaussianHair represent hair with Gaussians aligned to strand directions~\cite{zakharov2024haircut,luo2024gaussianhair}.
Our reconstruction optimizes rooted polylines without a learned strand prior and passes the resulting curves directly to position-based simulation~\cite{mueller2007pbd,mueller2012dftl}.
We use the public strand models from GeomHair \cite{lazuardi2026geomhair} as reference geometry~\cite{yuksel2026hairmodels}.

\vfill

\subsection{Volume rendering}
\label{sec:vr3dgs}
Given a density field $\sigma(\cdot)$ and an emitter color field $\bm c(\cdot)$, the color along a ray $\bm r(t)=\vo+t\vd$ as per volume rendering~\cite{max1995optical} can be computed as: 
\begin{equation}
  \bm C=\int_0^\infty T(t)\,\sigma(\bm r(t))\,\bm c(\bm r(t))\,\dd t,\\
  \label{eq:vre}
\end{equation}
where the transmittance term is defined as
\begin{equation}
  T(t)=\exp\!\left(-\int_0^t\sigma(\bm r(s))\,\dd s\right).
  \label{eq:transmittance}
\end{equation}
Splitting the ray into ordered intervals $I_k$ from front to back, with a constant color $\bm c_k$ on each, turns \cref{eq:vre} into front-to-back compositing,
\begin{equation}
  \bm C=\sum_{k=1}^{K}T_k\,\alpha_k\,\bm c_k,
  \label{eq:composite}
\end{equation}
where $\alpha_k$ is the opacity of interval $I_k$ and $T_k$ is the fraction of light that survives the intervals in front of it, given by
\begin{equation}
  \tau_k {=} \int_{I_k}\sigma(\bm r(t))\,\dd t,
  \quad
  \alpha_k{=}1-e^{-\tau_k},
  \quad
  T_k{=}\prod_{j<k}(1-\alpha_j).
  \nonumber
\end{equation}

\section{Method}
\label{sec:method}

We begin by reviewing how a 3D Gaussian can be interpreted as the convolution of a Dirac mass with an anisotropic Gaussian kernel~(\cref{sec:prelim}).
We then extend this formulation to Gaussian-convolved line segments~(\cref{sec:conv_seg}) and derive the corresponding closed-form rendering equation~(\cref{sec:exact}).
Finally, we describe fast rendering with EWA splatting~(\cref{sec:filter}), anti-aliasing~(\cref{sec:antialias}), and optimization~(\cref{sec:optimization}).

\subsection{3D Gaussians as a convolution}
\label{sec:prelim}
One of the 3D Gaussian primitives $\gaussian$ of \Cref{fig:bonsai_components}~(left) can be interpreted as the convolution of a Dirac mass placed at position $\vmu$ with a zero-mean Gaussian kernel $\calN_{\mS}{=}\calN(\bm 0, \mS)$~:
\begin{equation}
  \gaussian(\vx)
  =\big[\delta(\cdot-\vmu) \conv \calN_{\mS}\big](\vx).
  \label{eq:threedgs-density}
\end{equation}
Here, $\vx{\in}\mathbb R^3$ is a query point, $\delta$ is the three-dimensional Dirac distribution, $\mS$ is the covariance of the kernel, and $\conv$ denotes convolution.
Hence, $\gaussian$ is the Gaussian density with mean $\vmu$ and covariance $\mS$.
This interpretation decouples the support of the primitive from the kernel, which determines how the primitive extends through 3D space.
In the next section, we replace the point support with a segment and connect segments into polylines to obtain our \method curvilinear primitive; see~\Cref{fig:bonsai_components}~(right).

\subsection{\method}
\label{sec:conv_seg}
Following the interpretation in \Cref{sec:prelim}, we retain the kernel of 3DGS and replace the point support with a finite segment of learnable length.
A segment is defined by its endpoints $\vA, \vB \in \mathbb R^3$.
We write $\vv {=} \vB {-} \vA$ and parameterize the segment as $\vx(u) {=} \vA {+} u\vv$ with $u \in [0,1]$.
The segment is then represented by the measure
\begin{equation}
  m(\vx) = \int_0^1 \delta \big(\vx-\vx(u)\big) \,\dd u,
  \label{eq:segment-measure}
\end{equation}
Our primitive $\prim$ is the convolution of this segment measure with an anisotropic Gaussian kernel,
\begin{equation}
  \begin{aligned}
    \prim(\vx)&=\,[m\conv \calN_{\mS}](\vx).
  \end{aligned}
  \label{eq:primitive}
\end{equation}
When $\vA{=}\vB$, the above recovers the 3DGS point-kernel, hence \method is a \ul{strict} generalization of 3DGS representations.
The formulation in \cref{eq:primitive} extends naturally to other convolution kernels, such as the beta kernels of~\cite{liu2025dbs}.

\subsection{Closed-form volume rendering}
\label{sec:exact}
We now describe how our primitive defines a volumetric density field, and, without loss of generality, derive the closed-form volume rendering integral for the \textit{Gaussian} kernel.
We parameterize each primitive with a volumetric density that follows the support of its kernel.
For a point $\vx$, the density is
\begin{equation}
  \sigma(\vx)=\rho\,\big[m\conv \calN_{\mS}\big](\vx).
  \label{eq:sigma-primitive}
\end{equation}
Expanding the convolution into an integral over a point gives
\begin{equation}
  \sigma(\vx)=\rho\int m(\bm y)\,\calN_{\mS}(\vx-\bm y)\,\dd\bm y .
  \label{eq:sigma-conv}
\end{equation}
Substituting the segment measure of \cref{eq:segment-measure} and swapping the order of integration yields
\begin{equation}
  \sigma(\vx)=\rho\int_0^1\left[\int_{\bm y}\delta\big(\bm y-\vx(u)\big)\,\calN_{\mS}(\vx-\bm y)\,\dd\bm y\right]\dd u .
  \label{eq:sigma-swap}
\end{equation}
The inner integral over $\bm y$ collapses by the \textit{sifting} property of the Dirac delta, which results in
\begin{equation}
  \sigma(\vx)=\rho\int_0^1 \calN_{\mS}\big(\vx-\vx(u)\big)\,\dd u .
  \label{eq:sigma-sift}
\end{equation}
Intuitively, \cref{eq:sigma-sift} states that the density at $\vx$ is the kernel response accumulated over every point of the segment, which is precisely what convolving a segment with a Gaussian kernel should produce.
We now turn from the density to the opacity that a primitive contributes along a camera ray.
Following~\cite{tagliasacchi2022volumerenderingdigestfor}, the opacity along a ray $\bm r(t)=\vo+t\vd$ is
\begin{equation}
  \alpha(\bm r)=1-\exp\!\big(-\tau(\bm r)\big),
  \,
  \tau(\bm r)=\int_{-\infty}^{\infty}\sigma\big(\bm r(t)\big)\,\dd t .
  \label{eq:alpha-ray}
\end{equation}
Substituting the density from \cref{eq:sigma-sift}, evaluated at the points $\bm r(t)$ along the ray, we can write the $\tau(r)$ as
\begin{equation}
  \tau(\bm r)=\int_{-\infty}^{\infty}\rho\int_0^1 \calN_{\mS}\big(\bm r(t)-\vx(u)\big)\,\dd u\,\dd t .
  \label{eq:tau-double}
\end{equation}
Since $\calN_{\mS}$ is non-negative, Tonelli's theorem allows us to swap the order of integration, which gives
\begin{equation}
  \tau(\bm r)=\rho\int_0^1\int_{-\infty}^{\infty} \calN_{\mS}\big(\bm r(t)-\vx(u)\big)\,\dd t\,\dd u .
  \label{eq:tau-swap}
\end{equation}
The inner integral is the integral of a Gaussian along a ray, which is itself a one-dimensional Gaussian in the segment coordinate $u$.
Denoting this inner integral by $f(u)$, we simplify~\cref{eq:tau-swap} to
\begin{align}
  \tau(\bm r)&=\int_0^1\rho\,f(u)\,\dd u,
  \label{eq:tau-f}
  \\
  f(u)&=\int_{-\infty}^{\infty} \calN_{\mS}\big(\bm r(t)-\vx(u)\big)\,\dd t.
  \nonumber
\end{align}
Let us define the mean of the one-dimensional Gaussian $f(u)$ to be $\bar u$ and its variance to be $1/(2\kappa^2)$, then \cref{eq:tau-f} simplifies:
\begin{equation}
  \,f(u)=f(\bar u)\,e^{-\kappa^2(u-\bar u)^2}.
  \label{eq:f-gaussian}
\end{equation}
For a unit ray direction $\vd$, these quantities are
\begin{equation}
  \kappa^2=\lVert\vv\rVert^2_{\mQ_\perp},
  \qquad
  \bar u=\frac{\vv^\top\mQ_\perp(\vo-\vA)}{\kappa^2},
  \label{eq:uhat-kappa}
\end{equation}
where $\mQ{=}\tfrac12\mS^{-1}$ and $\mQ_\perp{=}\mQ-\mQ\vd\vd^\top\mQ/(\vd^\top\mQ\vd)$ is the matrix $\mQ$ of the marginal of $\calN_{\mS}$ along the camera ray.
Substituting \cref{eq:f-gaussian} into \cref{eq:tau-f} reduces the optical depth to a single integral,
\begin{equation}
  \begin{aligned}
    \tau(\bm r)&=\rho\,f(\bar u)\int_0^1 e^{-\kappa^2(u-\bar u)^2}\,\dd u,\\
    f(\bar u)&=\sqrt{\frac{\pi}{\vd^\top\mQ\vd}}\,e^{-\lVert\vo-\vx(\bar u)\rVert^2_{\mQ_\perp}}.
  \end{aligned}
  \label{eq:tau-finite}
\end{equation}
This integral has a closed form through the antiderivative of a Gaussian, $\int e^{-\kappa^2(u-\bar u)^2}\,\dd u=\tfrac{\sqrt\pi}{2\kappa}\operatorname{erf}\!\big(\kappa(u-\bar u)\big)$.
Evaluating it at the limits and using that $\operatorname{erf}$ is odd gives
\begin{equation}
  \begin{aligned}
    \tau(\bm r)
    &=\frac{\rho\,f(\bar u)\sqrt\pi}{2\kappa}
      \Big[\operatorname{erf}\!\big(\kappa\bar u\big)+\operatorname{erf}\!\big(\kappa(1-\bar u)\big)\Big].
  \end{aligned}
  \label{eq:tau-closed}
\end{equation}
Substituting \cref{eq:tau-closed} into \cref{eq:alpha-ray} yields the opacity of the primitive along the ray in closed form. The same pipeline also supports other kernel profiles, such as the compact Beta kernel of DBS~\cite{liu2025dbs}. The full derivation of the screen-space expressions is given in the \SupplementaryMaterial.

For polylines, we treat each segment independently: we evaluate the volume integral and the resulting opacity separately for every segment, and then composite the segments along the ray as individual primitives; see~\Cref{fig:convergence}. 

\begin{figure}[t]
\centering
\includegraphics[width=\linewidth]{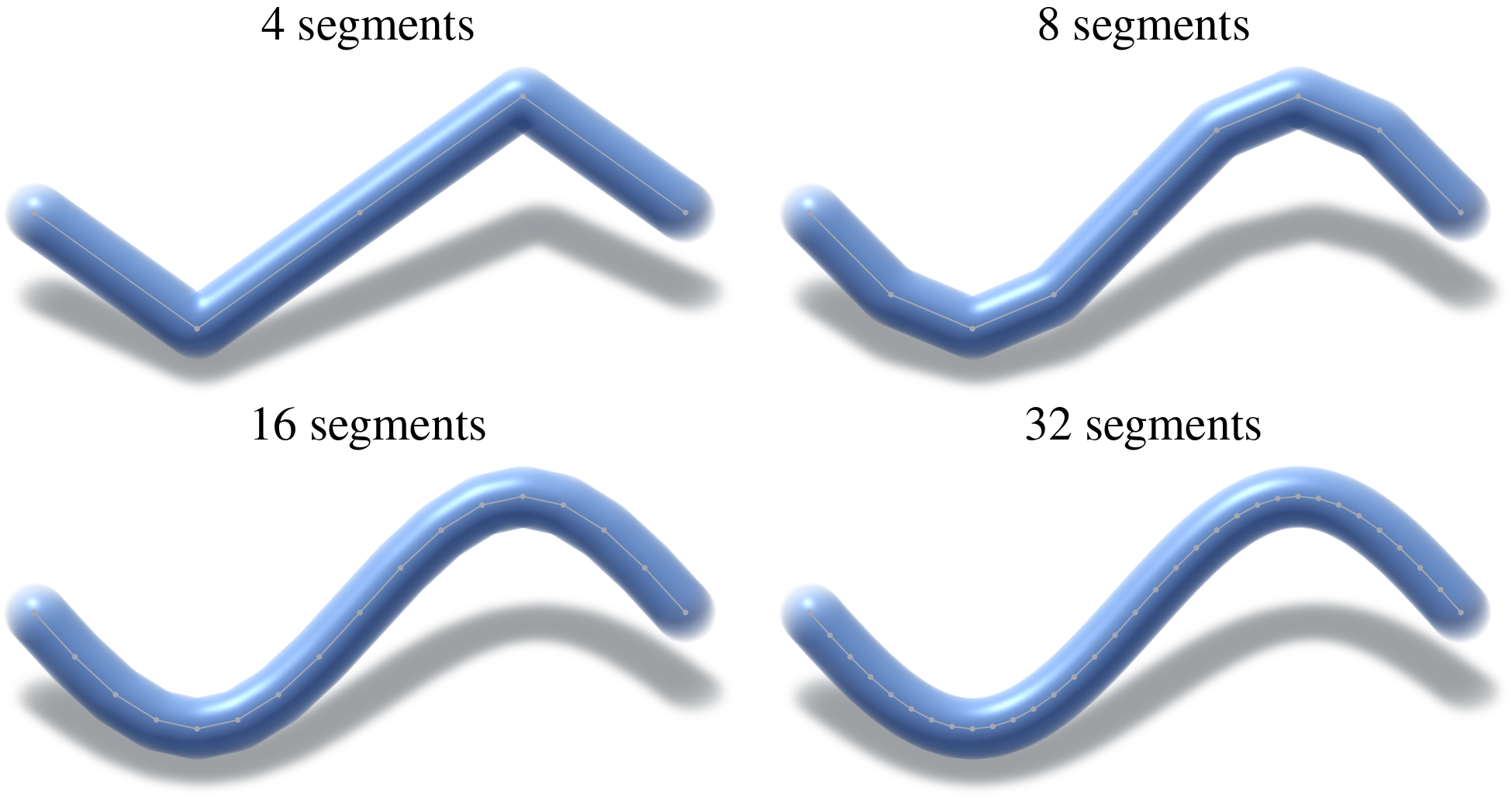}
\caption{
\textbf{Convergence} -- A curve rendered as a polyline of 4, 8, 16, and 32 Beta segments rasterized independently converges to the smooth curve without knots at the joints.
}
\label{fig:\currfilebase}\end{figure}

\subsection{Efficient rendering}
\label{sec:filter}
Evaluating the closed form for every primitive--pixel pair requires extensive computation of ray-dependent quantities, which is too slow for splatting-style training and rendering.
We therefore compute EWA~\cite{zwicker2002ewa} at the midpoint once per frame and scale it along the segment with inverse depth, which reduces the per-pixel evaluation to a few screen-space quantities.
This allows the kernel to vary in screen space, keeping the 3D shape consistent.
The yellow segment in \Cref{fig:method} shows the resulting foreshortening.
We additionally show analysis of our EWA computation in the \SupplementaryMaterial.

\subsection{Anti-aliasing}
\label{sec:antialias}
To avoid aliasing, we follow Mip-Splatting~\cite{yu2024mipsplatting} and apply a 3D as well as a 2D Gaussian filter to our primitives.
The 3D filter band-limits each primitive to the maximal sampling rate of the training views, so that it cannot encode detail the training images cannot resolve.
The 2D filter convolves each projected primitive with a sub-pixel screen-space Gaussian, which prevents primitives that project to less than a pixel from aliasing or flickering.
Under the EWA linearization~(\cref{sec:filter}), both filters simply add to the covariance of the projected kernel along the entire segment, so the closed form still applies.
We thus pre-filter each primitive in closed form, with a single sample per pixel.
In contrast, the concurrent work of \citet{tojo2026lines} \textit{does not} support spatial convolutions and must therefore resort to multisample anti-aliasing with four samples per pixel.

\subsection{Optimization}
\label{sec:optimization}
\begin{figure}[t]
\centering
\includegraphics[width=\linewidth]{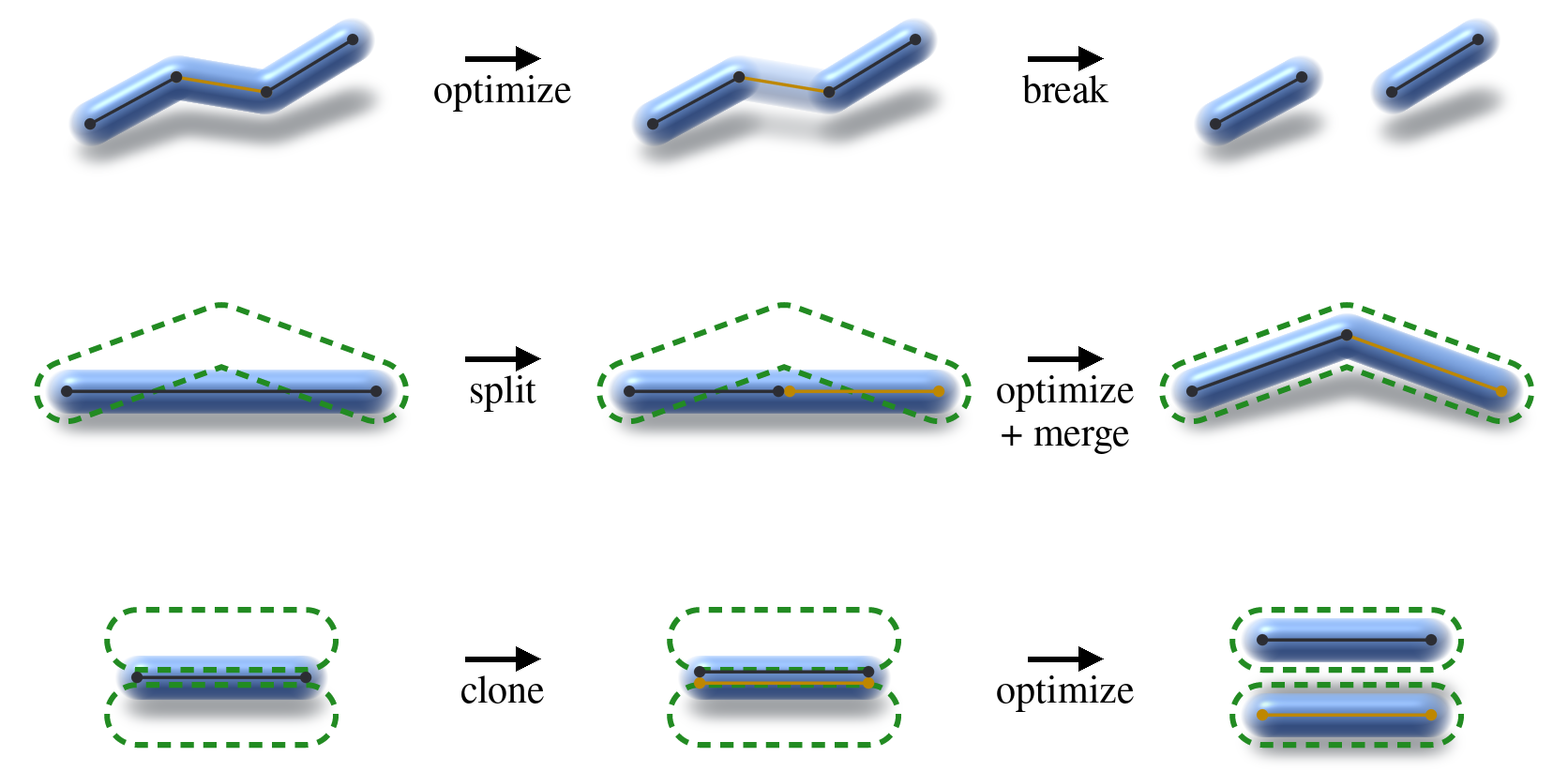}
\caption{
\textbf{Densification} --
We take inspiration from the densification strategy of 3DGS~MCMC~\cite{kheradmand2024mcmc}, but adapt it to our primitives.
When the opacity of a segment falls below a threshold, MCMC relocation moves it, potentially breaking up the polyline it was in.
The relocation target is then sampled in two ways. Long segments are split into two, short segments are cloned.
}
\label{fig:\currfilebase}\end{figure}

\begin{figure*}[t]
\centering
\includegraphics[width=\linewidth]{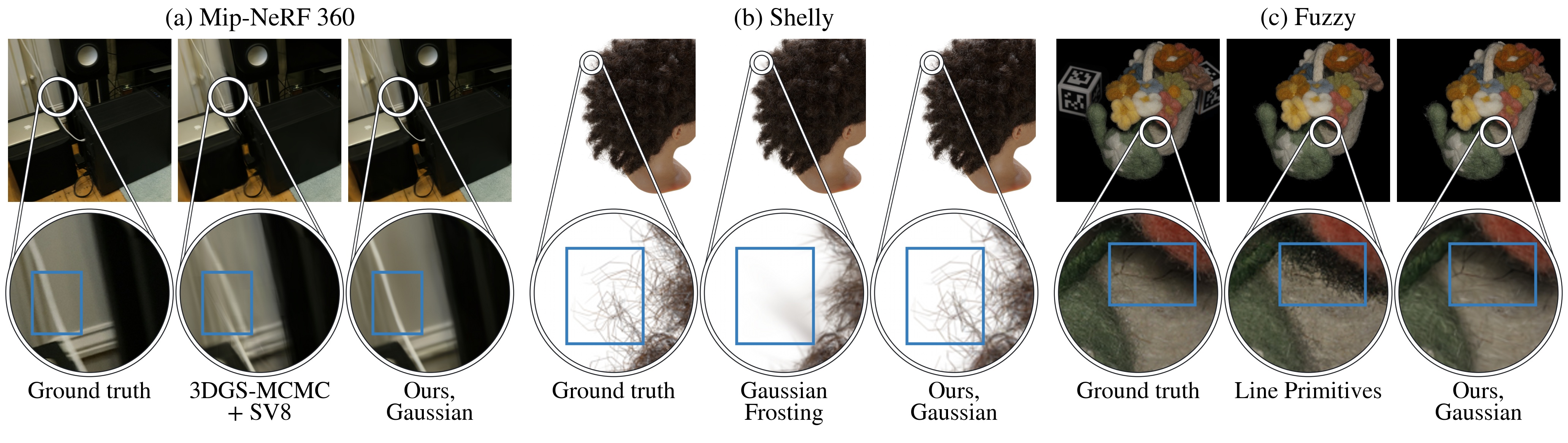}
\par\medskip
\footnotesize
\setlength{\tabcolsep}{4pt}
\resizebox{\linewidth}{!}{\begin{tabular}{l cccc cccc cccc}
& \multicolumn{4}{c}{Mip-NeRF~360~\cite{barron2022mipnerf360}} & \multicolumn{4}{c}{Shelly~\cite{wang2023shells}} & \multicolumn{4}{c}{Fuzzy~\cite{tojo2026lines}} \\
\cmidrule(lr){2-5} \cmidrule(lr){6-9} \cmidrule(lr){10-13}
Method & PSNR $\uparrow$ & SSIM $\uparrow$ & LPIPS $\downarrow$ & FPS $\uparrow$ & PSNR $\uparrow$ & SSIM $\uparrow$ & LPIPS $\downarrow$ & FPS $\uparrow$ & PSNR $\uparrow$ & SSIM $\uparrow$ & LPIPS $\downarrow$ & FPS $\uparrow$ \\
\midrule
Line Primitives~\cite{tojo2026lines} & 8.66 & 0.350 & 0.571 & \mgold{287} & 33.75 & 0.932 & 0.068 & \mgold{731} & 33.64 & 0.939 & 0.055 & \mgold{858} \\
Gaussian Frosting~\cite{guedon2024frosting} & 26.93 & 0.804 & 0.209 & 59 & 36.33 & 0.955 & \mgold{0.047} & 121 & 34.29 & 0.948 & \mbronze{0.043} & 124 \\
3DGS-MCMC + SV8~\cite{kheradmand2024mcmc,disario2026sv} & 28.39 & 0.834 & \msilver{0.190} & \msilver{198} & 37.10 & \msilver{0.959} & 0.050 & \msilver{464} & \msilver{35.00} & \msilver{0.956} & \msilver{0.041} & \msilver{422} \\
DBS + SV8~\cite{liu2025dbs,disario2026sv} & \msilver{28.71} & \msilver{0.837} & \mgold{0.189} & 125 & \msilver{37.22} & \msilver{0.959} & \mbronze{0.049} & 297 & \mbronze{34.98} & \msilver{0.956} & \mgold{0.040} & 283 \\
\midrule
Ours, Gaussian kernel & \mbronze{28.56} & \mbronze{0.835} & 0.200 & \mbronze{177} & \mgold{37.38} & \mgold{0.962} & \msilver{0.048} & \mbronze{452} & \mgold{35.03} & \mgold{0.957} & \mgold{0.040} & \mbronze{415} \\
Ours, Beta kernel & \mgold{28.78} & \mgold{0.838} & \mbronze{0.193} & 129 & \mbronze{37.12} & \mbronze{0.958} & \mbronze{0.049} & 284 & 34.90 & \mbronze{0.955} & \mgold{0.040} & 257 \\
\end{tabular}}
\captionof{figure}{\textbf{Comparison on Mip-NeRF~360, Shelly, and Fuzzy} -- 
(a, b) Unlike point-based primitives~\cite{kheradmand2024mcmc,disario2026sv,guedon2024frosting}, our method can reconstruct thin elongated geometry effectively.
(c) We can also model wide texture-less areas, like the dark shadow on the figure, with wide kernels, unlike Line Primitives~\cite{tojo2026lines}.
We mark the three best values in each column with gold, silver, and bronze medals.
}
\label{fig:\currfilebase}
\end{figure*}

\begin{figure}[t]
\centering
\includegraphics[width=\linewidth]{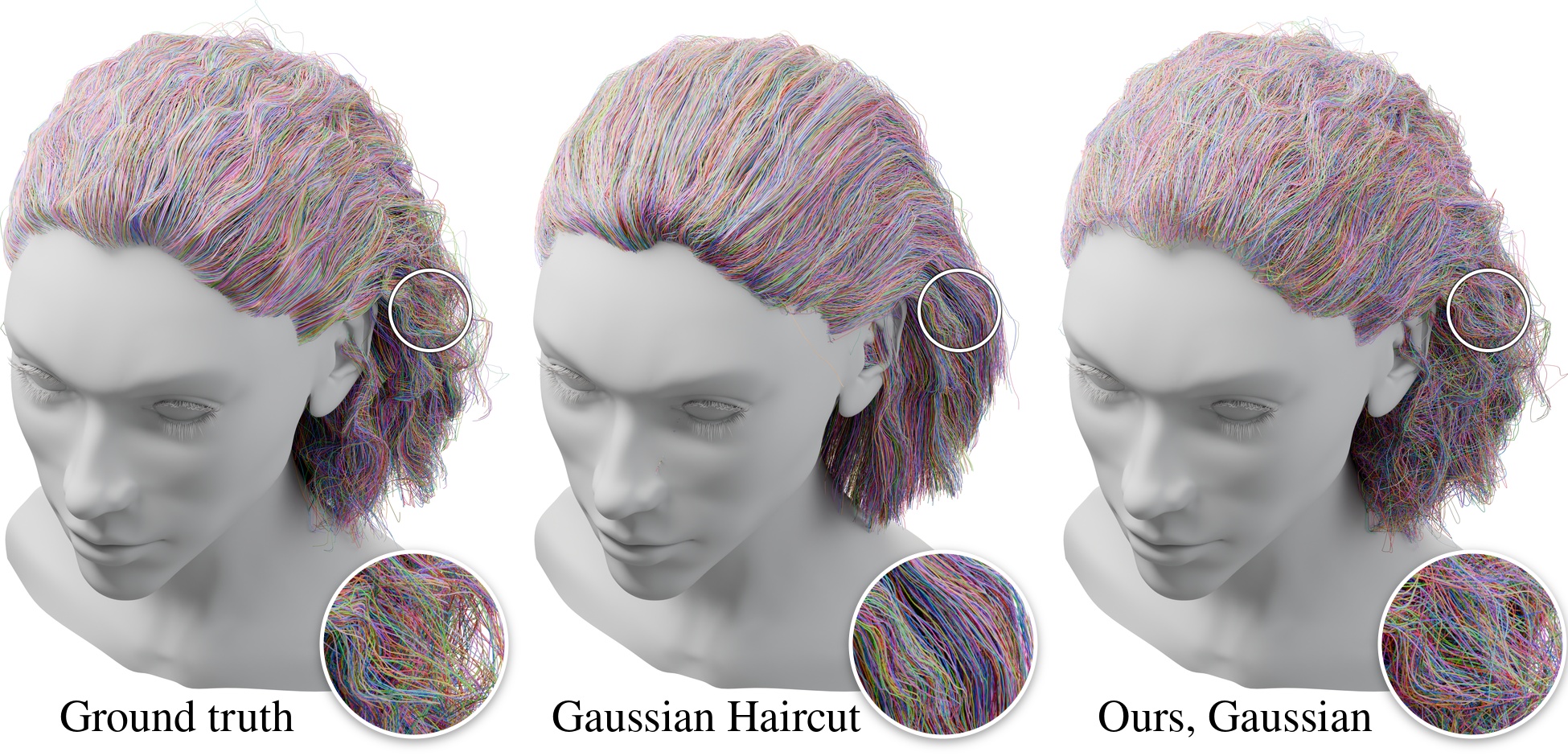}
\par\medskip
\footnotesize
\setlength{\tabcolsep}{8pt}
\resizebox{\linewidth}{!}{\begin{tabular}{l cc}
Method & PSNR $\uparrow$ / SSIM $\uparrow$ / LPIPS $\downarrow$ & F1 at 2 / 3 / 4\,mm, $\uparrow$ \\
\midrule
Gaussian Haircut~\cite{zakharov2024haircut}  & 29.56 / 0.923 / 0.064 & 46.9 / 63.5 / 72.7 \\
Ours, Gaussian Kernel & \textbf{32.25} / \textbf{0.940} / \textbf{0.043} &
\textbf{61.2} / \textbf{80.3} / \textbf{88.9} \\
\end{tabular}}
\captionof{figure}{\textbf{Comparison with Gaussian Haircut~\cite{zakharov2024haircut}} -- averaged over two datasets: GeomHair~\cite{lazuardi2026geomhair}~(selected 8 hairstyles) and the curly hair model of \citet{yuksel2026hairmodels}.
Top: strands of the curly hair model in a held-out view, one random color per strand.
Our strands closely match the ground-truth strands, whereas those of Gaussian Haircut are poorly structured.
}
\label{fig:\currfilebase}
\end{figure}

We optimize our representation with the objective
\begin{equation}
  \mathcal L=\mathcal L_{\mathrm{recon}}
  +\lambda_1 \mathcal L_{\mathrm{MCMC}}
  +\lambda_2 \mathcal L_{\mathrm{connect}},
  \label{eq:loss}
\end{equation}
where $\mathcal L_{\mathrm{recon}}$ is the standard photometric loss, a weighted combination of an $L_1$ and a D-SSIM term, between the rendered image $\hat{\bm I}$ and the ground-truth image $\bm I$.
The $\mathcal L_{\mathrm{MCMC}}$ term is the opacity and width regularization of~\cite{kheradmand2024mcmc}.
During optimization, we merge polylines whose endpoints fall within a predefined distance threshold.
To encourage this, $\mathcal L_{\mathrm{connect}}$ actively pulls nearby polyline endpoints together by minimizing the Euclidean distance between them.
We describe this attraction loss in more detail in the \SupplementaryMaterial.

\paragraph{Densification}
As in other per-scene optimization methods, we adaptively resample our primitives to allocate representation capacity where it is needed.
Following 3DGS-MCMC~\cite{kheradmand2024mcmc}, we prune segments and sample their relocation targets based on opacity.
We augment the MCMC relocation to additionally split long segments into two halves~(\cref{fig:densification}).
Notably, subdividing a segment leaves the rendering in image space \ul{exactly} unchanged when both children inherit the density of the parent; see the \SupplementaryMaterial.

\section{Experiments}
\label{sec:experiments}
We evaluate our primitive through extensive comparisons with a range of baselines on two tasks: novel view synthesis of general scenes~(\cref{sec:exp_scenes}) to demonstrate how this representation is a generalization of 3DGS, and reconstruction of hair strands from images~(\cref{sec:exp_hair}) to demonstrate its effectiveness its in-domain setting.

\subsection{General scenes}
\label{sec:exp_scenes}

We evaluate our method on $23$ scenes from three diverse datasets -- Mip-NeRF~360, Shelly and Fuzzy dataset.
Mip-NeRF~360~\cite{barron2022mipnerf360} is the standard benchmark for unbounded scenes, with five outdoor and four indoor scenes.
Shelly~\cite{wang2023shells} contains six synthetic objects with fur, wool, and thin leaves.
The very recently introduced Fuzzy dataset~\cite{tojo2026lines} contains eight captured objects with fibrous surfaces, such as cactus spines, fur, and brush bristles.
We follow the standard evaluation protocol of each dataset.
On Mip-NeRF~360, we downsample the indoor scenes by a factor of two and the outdoor scenes by a factor of four, and we hold out every eighth image for testing~\cite{kerbl2023gaussians}.
On Fuzzy, we downsample the images by a factor of four and likewise hold out every eighth view for testing.
On Shelly, we train and evaluate on images of $1920{\times}1080$ pixels with a white background.
The released Shelly train/test split \ul{mistakenly} contains test cameras that also appear in the training set.
We therefore remove from the training set all images that coincide with a test image.

\paragraph{Training}
We train for 30k iterations on all datasets, except for the outdoor scenes of Mip-NeRF~360, which, following Spherical Voronoi~(SV)~\cite{disario2026sv}, we train for 20k iterations.
MCMC relocation~\cite{kheradmand2024mcmc} stops 5k iterations before the end of training, and polyline merging~(\cref{sec:optimization}) runs only after relocation stops.
We cap the number of primitives at 500k on Shelly and Fuzzy, and at the per-scene budgets of Spherical Voronoi on Mip-NeRF~360.
All our models use the Spherical Voronoi appearance model with eight sites~(SV8)~\cite{disario2026sv}.

\paragraph{Baselines}
We compare with two point-based methods that use the SV8~\cite{disario2026sv} appearance model: 3DGS-MCMC~\cite{kheradmand2024mcmc} + SV8 and Deformable Beta Splatting~(DBS)~\cite{liu2025dbs} + SV8.
We also compare with Gaussian Frosting~\cite{guedon2024frosting} and with the \ul{concurrent} Line Primitives by~\citet{tojo2026lines}, which optimizes opaque line primitives.

\paragraph{Metrics}
On the test views, we report PSNR, SSIM~\cite{wang2004ssim}, and LPIPS~\cite{zhang2018lpips} (with a VGG backbone).
We also report the FPS at test resolution, measured on an RTX~4090 GPU.

\paragraph{Comparisons -- \Cref{fig:benchmarks}}
Our method with the Beta kernel outperforms all baselines on Mip-NeRF~360, and our method with the Gaussian kernel outperforms all baselines on Shelly and Fuzzy; see~\Cref{fig:benchmarks}.
Our method also matches the frame rate of 3DGS-MCMC across all three datasets. 
Since Line Primitives \cite{tojo2026lines} rasterizes thin lines of \textit{constant} width, it cannot model real-world scenes. 
We provide qualitative video comparisons with the baselines on the project webpage in the \SupplementaryMaterial.

\subsection{Hair reconstruction}
\label{sec:exp_hair}

For Hair reconstruction, we evaluate our method on a subset of GeomHair~\cite{lazuardi2026geomhair}.
GeomHair provides 400 strand-level hairstyles reconstructed from head scans, we select eight hairstyles that cover a wide range of hair length, curvature, and arrangement and place each hairstyle on the head mesh of GeomHair and render 100 views at $1200\times800$ pixels from cameras around the head, path-traced in Blender with the hair shading model of \citet{chiang2016hair}.
We hold out every eighth view, which leaves 88 training and 12 test views.
We additionally evaluate on the challenging curly hair model of \citet{yuksel2026hairmodels}, which we render and split in the same way. 
We report each metric as the average across the two datasets and show per-haircut breakdown in the \SupplementaryMaterial.

\paragraph{Training and baseline}
We train for 30k iterations, following the same protocol as for the general scenes.
We compare against Gaussian Haircut~\cite{zakharov2024haircut}, which relies on hair-specific priors.
By default, they prune all Gaussians outside a sphere fitted to the hair, which removes the chin and neck on short hairstyles.
We enlarge this sphere to enclose the head mesh, so that both methods render the entire head.

\paragraph{Metrics}
We report PSNR, SSIM, and LPIPS on the test views, rendered over a white background.
To evaluate the strand geometry, we compare our strands with the ground-truth strands of each hairstyle after resampling both at 0.5\,mm.
Following \citet{takimoto2024drhair}, a point in one strand set is considered correct if the other set contains a point within a distance threshold whose tangent line differs by less than an angle threshold.
We report the F1 score based on this criterion at the thresholds 2\,mm and $20^\circ$, 3\,mm and $30^\circ$, and 4\,mm and $40^\circ$~\cite{sklyarova2023neuralhaircut}.

\paragraph{Comparisons}
We show that our method outperforms Gaussian Haircut on all four metrics, with a 2.7\,dB higher PSNR and a 14.3-point higher F1 score at 2\,mm.
We show all the haircuts used in the \SupplementaryMaterial.
Like for the general scenes, we also provide qualitative video comparisons with the baselines on our project webpage.

\subsection{Ablation study}
We ablate the main design choices of our method to measure the effect of each one.
Specifically, we test our densification strategies~(midpoint splitting and MCMC-style segment cloning), polyline merging, and attraction regularization.
\Cref{tab:ablation_components} summarizes the quantitative effect of each choice.

\paragraph{Midpoint splits}
We assess the importance of midpoint splitting by removing it from the densification strategy and relying on segment cloning alone.
This slightly degrades performance, which shows that midpoint splitting contributes to training.

\paragraph{Segment cloning}
We assess segment cloning, following~\cite{kheradmand2024mcmc}, by disabling it and relying on midpoint splits alone.
This leads to a significant drop in reconstruction quality.
Without cloning, primitives cannot be effectively redistributed to regions that require short segments.

\paragraph{Polyline merging}
Next, we ablate the polyline merging strategy by disabling it during training.
This has no effect on the reconstruction metrics, since the representation can still model a connected polyline using many independent segments.
We can see this as the connected components become much smaller without merging, which leads to substantially less structural cohesiveness.

\paragraph{Attraction regularization}
We assess the attraction loss by disabling it during training.
This also has no direct effect on reconstruction quality.
However, it weakens polyline merging: without the attraction loss, strands are not encouraged to form a cohesive unit, which reduces the number of connected segments in a polyline.

\begin{table}[t]
\newcommand{\cmark}{\textcolor{dark_green}{\ding{51}}}
\newcommand{\xmark}{\textcolor{red}{\ding{55}}}
\centering
\small
\setlength{\tabcolsep}{3pt}
\resizebox{\linewidth}{!}{\begin{tabular}{cccc cc}
\shortstack{Midpoint\\split} & \shortstack{Segment\\cloning} & \shortstack{Polyline\\merging} & \shortstack{Attraction\\loss} & PSNR $\uparrow$ & \shortstack{Mean connected\\component size} \\
\midrule
\xmark & \cmark & \cmark & \cmark & 33.13 & 3.26 \\
\cmark & \xmark & \cmark & \cmark & 32.54 & 4.00 \\
\cmark & \cmark & \xmark & \cmark & 33.59 & 1.00 \\
\cmark & \cmark & \cmark & \xmark & 33.57 & 2.77 \\
\midrule
\cmark & \cmark & \cmark & \cmark & 33.54 & 3.50 \\
\end{tabular}}
\captionof{table}{\textbf{Ablations} -- Averaged over Fuzzy, Shelly, and Mip-NeRF~360; each row removes one component~(\xmark).
The size of a connected component is its number of segments; a segment without a joint has size 1.
}
\label{tab:\currfilebase}
\end{table}

\section{Conclusion}
\label{sec:conclusion}

We introduced a splatting representation formed by convolving finite segments with anisotropic kernels, with anisotropic point primitives recovered in the zero-length limit.
For Gaussian kernels, the contribution of a segment along a camera ray has a closed form, and the corresponding compositing rule allows connected segments to be rendered independently without changing the result under subdivision.
We also derived a screen-space approximation that supports compact profiles and renders at speeds comparable to point splatting.
The experiments show competitive image reconstruction and demonstrate that the same primitives can produce connected curves suitable for shading and simulation, while broader claims about automatically recovered topology require further evaluation.

\paragraph{Limitations and future work}
While our curvilinear primitives demonstrate a new capability, our representation also opens up many directions for future research.
Because high-frequency color requires high-frequency geometry, reproducing fine texture currently requires many short segments.
Combining the curvilinear primitives with learnable texture maps could reduce the number of primitives and improve the recovered geometry.
We have also not explored interpolating the kernel between the segment endpoints, which would extend our family of primitives to shapes such as cones, triangles, and trapezoids, and could further generalize the representation to surfaces.
Real hair reconstruction also remains challenging, and integrating learned hair priors more carefully is a promising direction.
Finally, it would be interesting to predict our representation with feedforward models, and to exploit the recovered curves to reverse engineer physical properties of the captured objects.

\paragraph{Acknowledgments}
George Shramko was supported in part by Natural Sciences and Engineering Research Council of Canada~(NSERC) CGRS-M scholarship.
This work was supported in part by the Natural Sciences and Engineering Research Council of Canada~(NSERC) Discovery Grant [2023-05617], NSERC Collaborative Research and Development Grant, the SFU Visual Computing Research Chair, Google Research, Digital Research Alliance of Canada, and Advanced Research Computing at the University of British Columbia.
We thank George Kopanas, Bernhard Kerbl, and Peter Hedman for their valuable feedback on early versions of the manuscript.
Their comments and suggestions helped us improve both the presentation and the clarity of this work.
We thank Will Eastcott and Dany Bittel for providing honeybee capture data.

\clearpage
{
    \small
    \bibliographystyle{ieeenat_fullname}
    \bibliography{refs}

\begin{thebibliography}{46}
\providecommand{\natexlab}[1]{#1}
\providecommand{\url}[1]{\texttt{#1}}
\expandafter\ifx\csname urlstyle\endcsname\relax
  \providecommand{\doi}[1]{doi: #1}\else
  \providecommand{\doi}{doi: \begingroup \urlstyle{rm}\Url}\fi

\bibitem[Barron et~al.(2021)Barron, Mildenhall, Tancik, Hedman, Martin-Brualla, and Srinivasan]{barron2021mipnerf}
Jonathan~T. Barron, Ben Mildenhall, Matthew Tancik, Peter Hedman, Ricardo Martin-Brualla, and Pratul~P. Srinivasan.
\newblock {Mip-NeRF}: A multiscale representation for anti-aliasing neural radiance fields.
\newblock In \emph{ICCV}, 2021.

\bibitem[Barron et~al.(2022)Barron, Mildenhall, Verbin, Srinivasan, and Hedman]{barron2022mipnerf360}
Jonathan~T. Barron, Ben Mildenhall, Dor Verbin, Pratul~P. Srinivasan, and Peter Hedman.
\newblock {Mip-NeRF 360}: Unbounded anti-aliased neural radiance fields.
\newblock In \emph{CVPR}, 2022.

\bibitem[Barron et~al.(2023)Barron, Mildenhall, Verbin, Srinivasan, and Hedman]{barron2023zipnerf}
Jonathan~T. Barron, Ben Mildenhall, Dor Verbin, Pratul~P. Srinivasan, and Peter Hedman.
\newblock {Zip-NeRF}: Anti-aliased grid-based neural radiance fields.
\newblock In \emph{ICCV}, 2023.

\bibitem[Bloomenthal and Shoemake(1991)]{bloomenthal1991convolution}
Jules Bloomenthal and Ken Shoemake.
\newblock Convolution surfaces.
\newblock In \emph{SIGGRAPH}, 1991.

\bibitem[Chiang et~al.(2016)Chiang, Bitterli, Tappan, and Burley]{chiang2016hair}
Matt Jen-Yuan Chiang, Benedikt Bitterli, Chuck Tappan, and Brent Burley.
\newblock A practical and controllable hair and fur model for production path tracing.
\newblock \emph{Computer Graphics Forum}, 35\penalty0 (2):\penalty0 275--283, 2016.

\bibitem[Choi et~al.(2024)Choi, Lee, Park, and Kim]{choi2024doodle}
Changwoon Choi, Jaeah Lee, Jaesik Park, and Young~Min Kim.
\newblock {3Doodle}: Compact abstraction of objects with {3D} strokes.
\newblock \emph{ACM Transactions on Graphics}, 43\penalty0 (4), 2024.

\bibitem[DiDonato et~al.(1980)DiDonato, Jarnagin, and Hageman]{didonato1980polygons}
A.~R. DiDonato, M.~P. Jarnagin, Jr., and R.~K. Hageman.
\newblock Computation of the integral of the bivariate normal distribution over convex polygons.
\newblock \emph{SIAM Journal on Scientific and Statistical Computing}, 1\penalty0 (2):\penalty0 179--186, 1980.

\bibitem[Gu{\'e}don and Lepetit(2024)]{guedon2024frosting}
Antoine Gu{\'e}don and Vincent Lepetit.
\newblock Gaussian frosting: Editable complex radiance fields with real-time rendering.
\newblock In \emph{European Conference on Computer Vision (ECCV)}, pages 413--430, 2024.

\bibitem[Held et~al.(2025)Held, Vandeghen, Hamdi, Deli{\`e}ge, Cioppa, Giancola, Vedaldi, Ghanem, and Droogenbroeck]{held2025convex}
Jan Held, Renaud Vandeghen, Abdullah Hamdi, Adrien Deli{\`e}ge, Anthony Cioppa, Silvio Giancola, Andrea Vedaldi, Bernard Ghanem, and Marc~Van Droogenbroeck.
\newblock {3D} convex splatting: Radiance field rendering with {3D} smooth convexes.
\newblock In \emph{CVPR}, 2025.

\bibitem[Held et~al.(2026{\natexlab{a}})Held, Son, Vandeghen, Rebain, Gadelha, Zhou, Cioppa, Lin, Droogenbroeck, and Tagliasacchi]{held2025meshsplatting}
Jan Held, Sanghyun Son, Renaud Vandeghen, Daniel Rebain, Matheus Gadelha, Yi Zhou, Anthony Cioppa, Ming~C. Lin, Marc~Van Droogenbroeck, and Andrea Tagliasacchi.
\newblock {MeshSplatting: Differentiable Rendering with Opaque Meshes}.
\newblock In \emph{Computer Vision and Pattern Recognition (CVPR)}, 2026{\natexlab{a}}.
\newblock oral.

\bibitem[Held et~al.(2026{\natexlab{b}})Held, Vandeghen, Deli{\`e}ge, Hamdi, Giancola, Cioppa, Vedaldi, Ghanem, Tagliasacchi, and Droogenbroeck]{held2026triangles}
Jan Held, Renaud Vandeghen, Adrien Deli{\`e}ge, Abdullah Hamdi, Silvio Giancola, Anthony Cioppa, Andrea Vedaldi, Bernard Ghanem, Andrea Tagliasacchi, and Marc~Van Droogenbroeck.
\newblock Triangle splatting for real-time radiance field rendering.
\newblock In \emph{3DV}, 2026{\natexlab{b}}.

\bibitem[Kazhdan et~al.(2006)Kazhdan, Bolitho, and Hoppe]{kazhdan2006poisson}
Michael Kazhdan, Matthew Bolitho, and Hugues Hoppe.
\newblock Poisson surface reconstruction.
\newblock In \emph{Symposium on Geometry Processing (SGP)}. The Eurographics Association, 2006.

\bibitem[Kerbl et~al.(2023)Kerbl, Kopanas, Leimk{\"u}hler, and Drettakis]{kerbl2023gaussians}
Bernhard Kerbl, Georgios Kopanas, Thomas Leimk{\"u}hler, and George Drettakis.
\newblock {3D} gaussian splatting for real-time radiance field rendering.
\newblock \emph{ACM Transactions on Graphics}, 42\penalty0 (4), 2023.

\bibitem[Kheradmand et~al.(2024)Kheradmand, Rebain, Sharma, Sun, Tseng, Isack, Kar, Tagliasacchi, and Yi]{kheradmand2024mcmc}
Shakiba Kheradmand, Daniel Rebain, Gopal Sharma, Weiwei Sun, Yang-Che Tseng, Hossam Isack, Abhishek Kar, Andrea Tagliasacchi, and Kwang~Moo Yi.
\newblock {3D} gaussian splatting as {Markov} chain {Monte Carlo}.
\newblock In \emph{NeurIPS}, 2024.

\bibitem[Lazuardi et~al.(2026)Lazuardi, Sevastopolsky, Zakharov, Niessner, and Sklyarova]{lazuardi2026geomhair}
Rachmadio~Noval Lazuardi, Artem Sevastopolsky, Egor Zakharov, Matthias Niessner, and Vanessa Sklyarova.
\newblock {GeomHair}: Reconstruction of hair strands from colorless {3D} scans, 2026.

\bibitem[Liu et~al.(2025)Liu, Sun, Chen, Wang, and Feng]{liu2025dbs}
Rong Liu, Dylan Sun, Meida Chen, Yue Wang, and Andrew Feng.
\newblock Deformable beta splatting.
\newblock In \emph{SIGGRAPH Conference Papers}, 2025.

\bibitem[Luo et~al.(2024)Luo, Ouyang, Zhao, Jiang, Zhang, Zhang, Yang, Xu, and Yu]{luo2024gaussianhair}
Haimin Luo, Min Ouyang, Zijun Zhao, Suyi Jiang, Longwen Zhang, Qixuan Zhang, Wei Yang, Lan Xu, and Jingyi Yu.
\newblock {GaussianHair}: Hair modeling and rendering with light-aware gaussians.
\newblock \emph{arXiv preprint arXiv:2402.10483}, 2024.

\bibitem[Mai et~al.(2025)Mai, Hedman, Kopanas, Verbin, Futschik, Xu, Kuester, Barron, and Zhang]{mai2025ever}
Alexander Mai, Peter Hedman, George Kopanas, Dor Verbin, David Futschik, Qiangeng Xu, Falko Kuester, Jonathan~T. Barron, and Yinda Zhang.
\newblock {EVER}: Exact volumetric ellipsoid rendering for real-time view synthesis.
\newblock In \emph{ICCV}, 2025.

\bibitem[Mallick et~al.(2024)Mallick, Goel, Kerbl, Steinberger, Carrasco, and {De La Torre}]{mallick2024taming}
Saswat~Subhajyoti Mallick, Rahul Goel, Bernhard Kerbl, Markus Steinberger, Francisco~Vicente Carrasco, and Fernando {De La Torre}.
\newblock Taming {3DGS}: High-quality radiance fields with limited resources.
\newblock In \emph{SIGGRAPH Asia 2024 Conference Papers}, pages 1--11. ACM, 2024.

\bibitem[Max(1995)]{max1995optical}
Nelson Max.
\newblock Optical models for direct volume rendering.
\newblock \emph{IEEE Transactions on Visualization and Computer Graphics}, 1\penalty0 (2):\penalty0 99--108, 1995.

\bibitem[Mildenhall et~al.(2020)Mildenhall, Srinivasan, Tancik, Barron, Ramamoorthi, and Ng]{mildenhall2020nerf}
Ben Mildenhall, Pratul~P. Srinivasan, Matthew Tancik, Jonathan~T. Barron, Ravi Ramamoorthi, and Ren Ng.
\newblock {NeRF}: Representing scenes as neural radiance fields for view synthesis.
\newblock In \emph{ECCV}, 2020.

\bibitem[Moenne-Loccoz et~al.(2024)Moenne-Loccoz, Mirzaei, Perel, de~Lutio, Esturo, State, Fidler, Sharp, and Gojcic]{moenneloccoz2024threedgrt}
Nicolas Moenne-Loccoz, Ashkan Mirzaei, Or Perel, Riccardo de Lutio, Janick~Martinez Esturo, Gavriel State, Sanja Fidler, Nicholas Sharp, and Zan Gojcic.
\newblock {3D} gaussian ray tracing: Fast tracing of particle scenes.
\newblock \emph{ACM Transactions on Graphics}, 43\penalty0 (6), 2024.

\bibitem[M{\"u}ller et~al.(2007)M{\"u}ller, Heidelberger, Hennix, and Ratcliff]{mueller2007pbd}
Matthias M{\"u}ller, Bruno Heidelberger, Marcus Hennix, and John Ratcliff.
\newblock Position based dynamics.
\newblock \emph{Journal of Visual Communication and Image Representation}, 18\penalty0 (2):\penalty0 109--118, 2007.

\bibitem[M{\"u}ller et~al.(2012)M{\"u}ller, Kim, and Chentanez]{mueller2012dftl}
Matthias M{\"u}ller, Tae-Yong Kim, and Nuttapong Chentanez.
\newblock Fast simulation of inextensible hair and fur.
\newblock In \emph{Workshop on Virtual Reality Interaction and Physical Simulation (VRIPHYS)}, 2012.

\bibitem[M{\"u}ller et~al.(2022)M{\"u}ller, Evans, Schied, and Keller]{mueller2022instant}
Thomas M{\"u}ller, Alex Evans, Christoph Schied, and Alexander Keller.
\newblock Instant neural graphics primitives with a multiresolution hash encoding.
\newblock \emph{ACM Transactions on Graphics}, 41\penalty0 (4), 2022.

\bibitem[Owen(1956)]{owen1956tables}
Donald~B. Owen.
\newblock Tables for computing bivariate normal probabilities.
\newblock \emph{The Annals of Mathematical Statistics}, 27\penalty0 (4):\penalty0 1075--1090, 1956.

\bibitem[Radl et~al.(2024)Radl, Steiner, Parger, Weinrauch, Kerbl, and Steinberger]{radl2024stopthepop}
Lukas Radl, Michael Steiner, Mathias Parger, Alexander Weinrauch, Bernhard Kerbl, and Markus Steinberger.
\newblock {StopThePop}: Sorted gaussian splatting for view-consistent real-time rendering.
\newblock \emph{ACM Transactions on Graphics}, 43\penalty0 (4), 2024.

\bibitem[Rosu et~al.(2022)Rosu, Saito, Wang, Wu, Behnke, and Nam]{rosu2022neuralstrands}
Radu~Alexandru Rosu, Shunsuke Saito, Ziyan Wang, Chenglei Wu, Sven Behnke, and Giljoo Nam.
\newblock Neural strands: Learning hair geometry and appearance from multi-view images.
\newblock In \emph{ECCV}, 2022.

\bibitem[Sario et~al.(2026)Sario, Rebain, Verbin, Grangetto, and Tagliasacchi]{disario2026sv}
Francesco~Di Sario, Daniel Rebain, Dor Verbin, Marco Grangetto, and Andrea Tagliasacchi.
\newblock Spherical {Voronoi}: Directional appearance as a differentiable partition of the sphere.
\newblock In \emph{CVPR}, 2026.

\bibitem[Sklyarova et~al.(2023)Sklyarova, Chelishev, Dogaru, Medvedev, Lempitsky, and Zakharov]{sklyarova2023neuralhaircut}
Vanessa Sklyarova, Jenya Chelishev, Andreea Dogaru, Igor Medvedev, Victor Lempitsky, and Egor Zakharov.
\newblock Neural haircut: Prior-guided strand-based hair reconstruction.
\newblock In \emph{ICCV}, 2023.

\bibitem[Tagliasacchi and Mildenhall(2022)]{tagliasacchi2022volumerenderingdigestfor}
Andrea Tagliasacchi and Ben Mildenhall.
\newblock Volume rendering digest (for nerf), 2022.

\bibitem[Takimoto et~al.(2024)Takimoto, Takehara, Sato, Zhu, and Zheng]{takimoto2024drhair}
Yusuke Takimoto, Hikari Takehara, Hiroyuki Sato, Zihao Zhu, and Bo Zheng.
\newblock {Dr.Hair}: Reconstructing scalp-connected hair strands without pre-training via differentiable rendering of line segments.
\newblock In \emph{CVPR}, 2024.

\bibitem[Talegaonkar et~al.(2025)Talegaonkar, Belhe, Ramamoorthi, and Antipa]{talegaonkar2025vol3dgs}
Chinmay Talegaonkar, Yash Belhe, Ravi Ramamoorthi, and Nicholas Antipa.
\newblock Volumetrically consistent {3D} gaussian rasterization.
\newblock In \emph{CVPR}, 2025.

\bibitem[Tojo et~al.(2026)Tojo, Shamir, Umetani, and Bickel]{tojo2026lines}
Kenji Tojo, Ariel Shamir, Nobuyuki Umetani, and Bernd Bickel.
\newblock Inverse rendering for modeling with line primitives.
\newblock \emph{ACM Transactions on Graphics}, 45\penalty0 (6), 2026.
\newblock SIGGRAPH Asia 2026.

\bibitem[Wang et~al.(2004)Wang, Bovik, Sheikh, and Simoncelli]{wang2004ssim}
Zhou Wang, Alan~C. Bovik, Hamid~R. Sheikh, and Eero~P. Simoncelli.
\newblock Image quality assessment: From error visibility to structural similarity.
\newblock \emph{IEEE Transactions on Image Processing}, 13\penalty0 (4):\penalty0 600--612, 2004.

\bibitem[Wang et~al.(2023)Wang, Shen, Nimier-David, Sharp, Gao, Keller, Fidler, M{\"u}ller, and Gojcic]{wang2023shells}
Zian Wang, Tianchang Shen, Merlin Nimier-David, Nicholas Sharp, Jun Gao, Alexander Keller, Sanja Fidler, Thomas M{\"u}ller, and Zan Gojcic.
\newblock Adaptive shells for efficient neural radiance field rendering.
\newblock \emph{ACM Transactions on Graphics}, 42\penalty0 (6), 2023.

\bibitem[Westover(1990)]{westover1990footprint}
Lee Westover.
\newblock Footprint evaluation for volume rendering.
\newblock In \emph{SIGGRAPH}, 1990.

\bibitem[Wu et~al.(2025)Wu, Esturo, Mirzaei, Moenne-Loccoz, and Gojcic]{wu2025threedgut}
Qi Wu, Janick~Martinez Esturo, Ashkan Mirzaei, Nicolas Moenne-Loccoz, and Zan Gojcic.
\newblock {3DGUT}: Enabling distorted cameras and secondary rays in gaussian splatting.
\newblock In \emph{CVPR}, 2025.

\bibitem[Xu et~al.(2025)Xu, Bitterli, d'Eon, and Jarosz]{xu2025gpis}
Kehan Xu, Benedikt Bitterli, Eugene d'Eon, and Wojciech Jarosz.
\newblock Practical {Gaussian} process implicit surfaces with sparse convolutions.
\newblock \emph{ACM Transactions on Graphics}, 44\penalty0 (6), 2025.

\bibitem[Ying and Zwicker(2025)]{ying2025sketchsplat}
Haiyang Ying and Matthias Zwicker.
\newblock {SketchSplat}: {3D} edge reconstruction via differentiable multi-view sketch splatting.
\newblock In \emph{ICCV}, pages 25649--25659, 2025.

\bibitem[Yu et~al.(2024)Yu, Chen, Huang, Sattler, and Geiger]{yu2024mipsplatting}
Zehao Yu, Anpei Chen, Binbin Huang, Torsten Sattler, and Andreas Geiger.
\newblock {Mip-Splatting}: Alias-free {3D} gaussian splatting.
\newblock In \emph{CVPR}, 2024.

\bibitem[Yuksel(2026)]{yuksel2026hairmodels}
Cem Yuksel.
\newblock {HAIR} model files.
\newblock \url{http://www.cemyuksel.com/research/hairmodels/}, 2026.

\bibitem[Zakharov et~al.(2024)Zakharov, Sklyarova, Black, Nam, Thies, and Hilliges]{zakharov2024haircut}
Egor Zakharov, Vanessa Sklyarova, Michael~J. Black, Giljoo Nam, Justus Thies, and Otmar Hilliges.
\newblock Human hair reconstruction with strand-aligned {3D} gaussians.
\newblock In \emph{ECCV}, 2024.

\bibitem[Zhang et~al.(2018)Zhang, Isola, Efros, Shechtman, and Wang]{zhang2018lpips}
Richard Zhang, Phillip Isola, Alexei~A. Efros, Eli Shechtman, and Oliver Wang.
\newblock The unreasonable effectiveness of deep features as a perceptual metric.
\newblock In \emph{CVPR}, 2018.

\bibitem[Zhu et~al.(2025)Zhu, Yue, He, and Wang]{zhu2025sss}
Jialin Zhu, Jiangbei Yue, Feixiang He, and He Wang.
\newblock {3D} student splatting and scooping.
\newblock In \emph{CVPR}, 2025.

\bibitem[Zwicker et~al.(2002)Zwicker, Pfister, van Baar, and Gross]{zwicker2002ewa}
Matthias Zwicker, Hanspeter Pfister, Jeroen van Baar, and Markus Gross.
\newblock {EWA} splatting.
\newblock \emph{IEEE Transactions on Visualization and Computer Graphics}, 8\penalty0 (3):\penalty0 223--238, 2002.

\end{thebibliography}
}

\end{document}